\documentclass{kerauth}
\usepackage{url}
\usepackage{natbib}
\usepackage{listings}
\usepackage{tabularx}
\usepackage{graphicx}

\newcommand{\MathWord}[1]{\mbox{\it #1}}

\begin{document}


\doublespacing

\title{Reformulation Techniques for Automated Planning:\\ A Systematic Review}

\author{DIAEDDIN ALARNAOUTI and GEORGE BARYANNIS and MAURO VALLATI}

\address{School of Computing \& Engineering, University of Huddersfield, HD1 3DH, Huddersfield, UK\\
\email{\{diaeddin.alarnaouti,g.bargiannis,m.vallati\}@hud.ac.uk}}

\begin{abstract}

Automated planning is a prominent area of Artificial Intelligence, and an important component for intelligent autonomous agents. A cornerstone of domain-independent planning is the separation between planning logic, i.e. the automated reasoning side, and the knowledge model, that encodes a formal representation of domain knowledge needed to reason upon a given problem to synthesise a solution plan. Such a separation enables the use of reformulation techniques, which transform how a model is represented in order to improve the efficiency of plan generation. Over the past decades, significant research effort has been devoted to the design of reformulation techniques. In this paper, we present a systematic review of the large body of work on reformulation techniques for classical planning, aiming to provide a holistic view of the field and to foster future research in the area. As a tangible outcome, we provide a qualitative comparison of the existing classes of techniques, that can help researchers gain an overview of their strengths and weaknesses. 



\end{abstract}

\section{Introduction}

Automated planning is a research discipline that addresses the problem of generating a totally- or partially-ordered sequence of actions that transform the environment from some initial state to a desired goal state. Within this discipline, domain-independent planning refers to those approaches that keep the knowledge model, the domain knowledge related to the problem at hand, separate from planning logic, that enables automated reasoning to generate plans. The development of domain-independent planners within the AI Planning community facilitates the use of this ``off-the-shelf'' technology for a wide range of applications, including UAV manoeuvring \citep{DBLP:conf/atal/RamirezPLBMPSZ18}, space exploration \citep{ai2004mapgen}, and train dispatching \citep{cardellini2021station}. This is despite the complexity issues inherent in plan generation, which are exacerbated by the separation of planner logic from domain knowledge. On the other hand, this separation has the advantage that planning engines can be interchanged in a modular way, provided that they accept the same language for describing planning problems and deliver the same type of plans.

This modular approach has fostered the development of planning engines, as well as the design and exploitation of reformulation techniques. These refer to the ability to automatically re-formulate, re-represent or tune the domain model and/or a problem description, while keeping to the same input language, in order to increase the efficiency of a planning engine and expand the scope of problems solved \citep{riddle2011does}. The aim is to make these techniques independent of domain and planner to some degree (that is, applicable to a range of domains and planning engine technology), and use them to form a wrapper around a planner, improving its overall performance for the particular domain to which it is applied.

In the past few decades, a number of reformulation techniques have been introduced. This is particularly true for classical planning which, despite its intrinsic simplicity, provides an ideal ground to study and investigate planning techniques that can then be extended beyond a specific framework. For this reason, classical planning has been studied for several decades, and is still a focal point of research within the automated planning community, as evidenced by the papers devoted to it in the flagship conference, the International Conference on Automated Planning and Scheduling (ICAPS)\footnote{The interested reader is referred to the ICAPS website: https://www.icaps-conference.org/} and by the number of papers focusing on knowledge engineering for classical planning that are accepted at the Workshop on Knowledge Engineering for Planning and Scheduling (KEPS).  

In this paper, we systematically review the state of the art of reformulation techniques for classical planning, with the aim of providing a holistic view of the field and answering the following research questions:

\begin{itemize}
    \item What reformulation techniques have been proposed in literature?
    \item How can they be applied on an indicative planning problem?
    \item What are the particular strengths and weaknesses of each individual reformulation technique?
\end{itemize}

The main purpose of this review is to provide a qualitative comparison of the existing reformulation approaches, with the aim of helping experts and practitioners in identifying the most promising technique to be used, and to highlight research gaps that can foster further research in the area.  Note that empirical analysis of the reviewed reformulation techniques is beyond the scope of this work due to the high variability, among different papers, of considered benchmarks, planning engines, and software/hardware infrastructure \citep{DBLP:journals/aicom/BoccheseFVGH18}. 

The remainder of this paper is organised as follows. Section~\ref{sec:background} the necessary background on planning and reformulation, and we introduce an example domain model to be used as a running example throughout the paper. Then, Section~\ref{sec:methodology} describes the methodology used for this literature review. Section~\ref{sec:ref} is the main part of the review, presenting the existing reformulation techniques, and showing how they can be implemented on the running example. We then compare the reviewed techniques in Section~\ref{sec:qual}, while in Section~\ref{sec:non} we briefly discuss some reformulation techniques that target non-classical planning problems. Finally, we provide conclusions and suggested directions for future research in the area of reformulation techniques for automated planning. 


\section{Background}
\label{sec:background}
This section is devoted to providing the required background in terms of automated classical planning, the Gripper domain model used as a running example, and the notion of reformulation. 

\subsection{Classical Planning}

Classical planning is concerned with finding a (partially or totally ordered) sequence of actions transforming the static, deterministic and fully observable environment from the given initial state to a desired goal state~\citep{apl}.

In the classical representation, a \textit{planning task} consists of a \textit{planning domain model} and a \textit{planning problem}, where the planning domain model describes the environment and defines planning operators while the planning problem defines concrete objects, an initial state and a set of goals. The environment is described by \textit{predicates} that are specified via a unique identifier and terms (variable symbols or constants). 

Formally, a \textit{planning task} is a pair $\Pi = (Dom_\Pi,Prob_\Pi)$ where a \textit{planning domain model} $Dom_\Pi = (P_\Pi,Ops_\Pi)$ is a pair consisting of a finite set of predicates $P_\Pi$ and planning operators $Ops_\Pi$, and a \textit{planning problem} $Prob_\Pi=(Objs_\Pi,I_\Pi,G_\Pi)$ is a triple consisting of a finite set of objects $Objs_\Pi$, initial state $I_\Pi$ and goal $G_\Pi$.

Let $ats_\Pi$ be the set of all \textit{atoms} that are formed from the predicates $P_\Pi$ by applying all possible substitution mappings from the predicate parameters (variable symbols) to the objects from $Objs_\Pi$. In other words, an atom is an \textit{instance} of a predicate (in this article, when we use the term instance, we mean an instance that is fully {\it ground}). A \textit{state} is a subset of $ats_\Pi$, and the \textit{initial state} $I_\Pi$ is a distinguished state. The \textit{goal} $G_\Pi$ is a non-empty subset of $ats_\Pi$, and a \textit{goal state} is any state that contains the goal $G_\Pi$. Note that the semantics of \textit{state} reflect the full observability of the environment; that is, for a state $s$, atoms present in $s$ are assumed to be true in $s$, while atoms not present in $s$ are assumed to be false in $s$.

\textit{Planning operators} are ``modifiers'' of the environment. They consist of \textit{preconditions}, i.e., what must hold prior to an operator's application, and \textit{effects}, i.e., what is changed after its application. We distinguish between \textit{negative effects}, i.e., what becomes false, and \textit{positive effects}, i.e., what becomes true after an operator's application. \textit{Actions} are instances of planning operators, i.e., an operator's parameters, as well as corresponding variable symbols in its preconditions and effects, are substituted by objects (constants). Planning operators capture general types of activities that can be performed. While predicates can be instantiated to atoms to capture given relations between concrete objects, planning operators can be instantiated to actions to capture given activities between concrete objects.


A \textit{planning operator}
$o=(\MathWord{name}(o),\MathWord{pre}(o),\MathWord{eff}(o))$
is specified such that $\MathWord{name}(o) =
\MathWord{op\_name}(x_1, \dots, x_k)$, where $\MathWord{op\_name}$ is a
unique identifier and $x_1, \dots, x_k$ are all the variable symbols
(parameters) appearing in the operator,
$\MathWord{pre}(o)$ is a set of predicates representing its precondition, and $\MathWord{eff}(o)$ represents its effects, divided into $\MathWord{eff}^-(o)$ and $\MathWord{eff}^+(o)$ (i.e., $\MathWord{eff}(o)=\MathWord{eff}^-(o)\cup \MathWord{eff}^+(o)$) that are sets of predicates representing the operator's negative and positive effects, respectively. \textit{Actions} are instances of planning operators that are formed by substituting objects, which are defined in a planning problem, for operators' parameters as well as for the corresponding variable symbols in operators' preconditions and effects. An action $a=(\MathWord{pre}(a),\MathWord{eff}^-(a)\cup\MathWord{eff}^+(a))$ is \textit{applicable} in a state $s$ if and only if $\MathWord{pre}(a)\subseteq s$. Application of $a$ in $s$, if possible, results in a state $(s\setminus\MathWord{eff}^-(a))\cup \MathWord{eff}^+(a)$.

A solution of a planning task is a sequence of actions transforming the environment from the given initial state to a goal state. A \textit{plan} is a sequence of actions. A plan is a \textit{solution} of a planning task $\Pi$, a \textit{solution plan} of $\Pi$ in other words, if and only if a consecutive application of the actions from the plan starting in the initial state of $\Pi$ results in the goal state of $\Pi$.

The standardised language for describing classical planning tasks is PDDL \citep{mcdermott20001998}, that was introduced in 1998 by the organisers of the first International Planning Competition,\footnote{https://www.icaps-conference.org/competitions/} building on top of STRIPS \citep{fikes1971strips} and the Action Description Language (ADL) \citep{pednault1987formulating}.

\subsection{The Gripper Domain}
As a running example in this paper, we consider the well-known Gripper domain model, initially introduced in the first International Planning Competition in 1998 by Jana Kohler \citep{mcdermott20001998}. This domain was selected in this paper because of its simplicity, and due to its suitability for being reformulated. In a nutshell, the Gripper domain consists of a robot that has two grippers, and is tasked to moved a number of balls between two rooms. The domain model includes three operators:
\begin{description}
   
   \item[Move:]  to move the robot between rooms.
   \item[Pick:] to use a gripper to pick up a ball.
   \item[Drop:] to drop a ball that the robot is holding in one of its grippers. 
\end{description}

Figure \ref{fig:gripper_domain} shows the PDDL code of the domain model, including the relevant predicates and the three mentioned operators. Figure \ref{fig:gripper_problem} shows an example Gripper planning problem, where the robot has 2 grippers, and has to move 4 balls between the considered rooms. The initial state defines the initial position of balls and robot, and the initial status of the grippers. The goal section specifies the desired goal position of the 4 balls. 

\begin{figure}
\small
\begin{verbatim}
(:predicates (room ?r) (ball ?b) (gripper ?g) (at-robby ?r) (at ?b ?r) (free ?g) (carry ?o ?g))
	
(:action move
       :parameters  (?from ?to)
       :precondition (and  (room ?from) (room ?to) (at-robby ?from))
       :effect (and  (at-robby ?to) (not (at-robby ?from))))

(:action pick
       :parameters (?obj ?room ?gripper)
       :precondition  (and  (ball ?obj) (room ?room) (gripper ?gripper) (at ?obj ?room) 
       										(at-robby ?room) (free ?gripper))
       :effect (and (carry ?obj ?gripper) (not (at ?obj ?room) (not (free ?gripper))))

(:action drop
       :parameters  (?obj  ?room ?gripper)
       :precondition  (and  (ball ?obj) (room ?room) (gripper ?gripper) 
       (carry ?obj ?gripper) (at-robby ?room))
       :effect (and (at ?obj ?room) (free ?gripper)  (not (carry ?obj ?gripper))))
\end{verbatim}
\caption{The Gripper PDDL domain model.}\label{fig:gripper_domain}
\end{figure}

\begin{figure}
\small
\begin{verbatim}
   (:objects rooma roomb ball4 ball3 ball2 ball1 left right)
   (:init (room rooma) (room roomb)
          (ball ball4) (ball ball3) (ball ball2) (ball ball1)
          (at-robby rooma)
          (free left) (free right)
          (at ball4 rooma) (at ball3 rooma) (at ball2 rooma) (at ball1 rooma)
          (gripper left) (gripper right))
   (:goal (and (at ball4 roomb) (at ball3 roomb) (at ball2 roomb) (at ball1 roomb)))
\end{verbatim}   
       \caption{An example Gripper planning problem that consists of 4 balls to be carried from \texttt{rooma} to \texttt{roomb}.}
    \label{fig:gripper_problem}
\end{figure}

\subsection{Reformulation in Domain-Independent planning}

Taking a general perspective, reformulation is a term meaning a change to the way in which one thinks about a problem, and it has been demonstrated to be a common practice employed also by people to tackle challenging problems \citep{riddle2013problem}. A theoretical framework for describing reformulation schemes in automated planning has been presented by~\cite{chrpa2012reformulating}. Early research in the area of reformulation in AI started in the 1960s \citep{Amarel1968OnRO}, and reformulation has rapidly become a major field of research. In the field of automated planning and, more general, of state-space search, reformulation is intended as a change of representation. Different representations of the same problem can result in different search spaces, and the use of reformulation techniques can allow to make a problem more amenable for a considered planning engine by providing a search space that is easier to be explored to find a goal state. 

Focusing on automated planning, the domain-independent paradigm decouples reasoning from knowledge representation. This supports the use of reformulation techniques which can re-formulate, re-represent or tune the domain model and/or problem description, while keeping to the same input language, in order to increase the efficiency of a planning engine and expand the scope of problems solved. The idea is that these techniques are independent from application domain and planning engines (that is, applicable to a range of domains and planning engine technology), and use them to form a wrapper around a planner, improving its overall performance for the domain to which it is applied.


\begin{figure}[t]
\includegraphics[width=.9\textwidth]{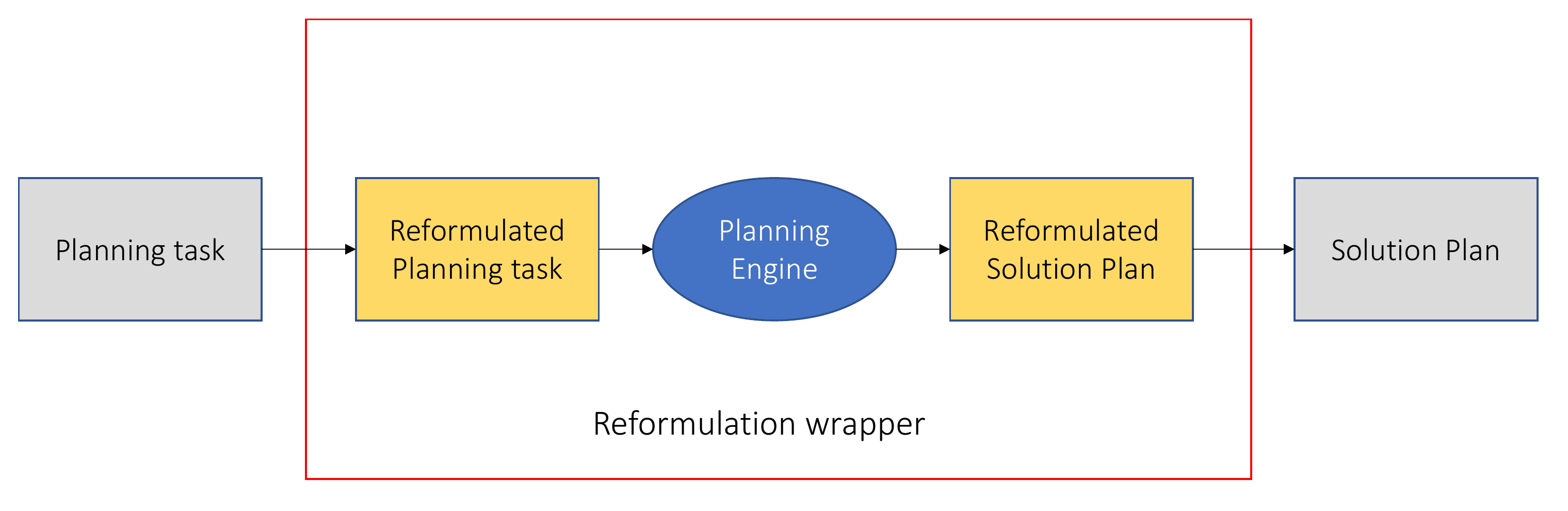}
        \caption{ \label{fig:domain_independent} An overview of the use of a domain- and planner-independent reformulation technique.}
    \end{figure}

Figure \ref{fig:domain_independent} gives an overview of the use of reformulation wrapper to modify the representation of a planning task in a domain- and planner-independent way that allows to reuse the same technique in any domain and with any engine that supports the considered formal language. In this review, we focus on classical planning reformulation approaches that aim at changing the representation of a planning task within the same formal language, with the goal of making the task more amenable for a given (class of) engines. 

       


\section{Review Methodology}
\label{sec:methodology}
This section is devoted to presenting the methodology used for performing the literature review. To ensure deeper understanding of the reformulation approaches that are applicable in domain-independent classical planning, we applied a variation of the systematic literature review methodology as proposed by~\cite{Kitchenham2017} for software engineering and previously applied for reviews of AI-related literature~\citep{Baryannis2019}. 

\subsection{Search strategy}\label{sec:strat}
The collection of literature concerning reformulation techniques in classical planning was performed through automated search using Google Scholar, Scopus and the University of Huddersfield library search engine (Summon). Two levels of keyword terms were utilised. On the first level, we used a disjunction (OR) of the following three keyword phrases: ``classical planning reformulation'', ``PDDL reformulation'', and ``domain-independent reformulation''. On the basis of the techniques identified using the aforementioned keyword phrases, we provided a second level of keywords, to obtain all the relevant works for each of them. This second-level search was with the following keywords linked to one another by OR: ``macro-operators'', ``macro-actions'', ``entanglements'', ``actions elimination'', ``bagged representation, ``action schema splitting'', and ``model configuration''.  

The previously mentioned libraries were used equally as results vary between them. The covered time period is set from 1980 to January 2022. Notably, thanks to the introduction of the standardised PDDL language and the first edition of the International Planning Competition in 1998, most of the relevant work has been published after 2000. 



To maximise coverage and completeness of the conducted search, two ancillary procedures were included: (1) checking reference lists of select primary studies (often referred to as backwards snowballing); and (2) identifying existing literature reviews on automated planning that may include reformulation (more details in Section~\ref{sec:relrev}).

\subsection{Search scope}
A series of inclusion and exclusion criteria frames the scope of this review, to ensure quality and relevance of the selected works. First, studies must be peer-reviewed and written in English. 
Then, they should contain a formal description of at least one reformulation technique for automated planning models, accompanied with an empirical analysis of the usefulness of the proposed technique. Third, the proposed reformulation technique must be domain-independent and must focus on classical planning. Fourth, the reformulation must accept planning models as input, and provide as output models that can be processed by domain-independent classical planning engines. It should be noted that the last point excludes some well-known approaches such as DISCOPLAN \citep{gerevini1998inferring} and TIM \citep{DBLP:journals/jair/FoxL98}, that do not allow to include the extracted knowledge in planning models. While the main focus of the review is on techniques for reformulating classical PDDL lifted models, here we also include works proposed before the introduction of the language, and that propose techniques for reformulating STRIPS \citep{fikes1971strips} models; for the purposes of this review these can be considered to be similar in nature to PDDL ones.  

A search based on the keywords and the engines described in Section \ref{sec:strat} yields several thousands of results. By carefully and thoroughly applying all aforementioned criteria and ancillary search strategies, and by considering also abstracts of the identified studies, $54$ studies remain and are analysed in the rest of this paper. For the sake of readability, the selected studies are divided into different classes, presented in Section \ref{sec:ref}, according to the reformulation approach that they present. 

\subsection{Related surveys}~\label{sec:relrev}
The main motivation behind this systematic review is the limited amount of works covering the field of domain-independent reformulation for automated planning. The most relevant published work is a survey of machine learning methods for automated planning \citep{jimenez2012review} where a section is dedicated to macro-actions. This survey covers only 4 studies that are also covered in this paper, published between 1977 and 2007. The work of~\cite{long2002reformulation} provides a definition of reformulation for automated planning, and briefly discusses some ways in which it has been applied in the field, though it also includes cases not within the scope of this review. To the best of our knowledge, this review is the first attempt at systematically analysing reformulation approaches for automated planning.


\section{Reformulation Techniques for Classical Planning}\label{sec:ref}

In this section we present the reviewed studies, classified according to the implemented reformulation approach. For each approach, we also provide an example of applying the reformulation approach to the Gripper domain.

\subsection{Entanglements} \label{Ent}

The entanglements approach was firstly introduced in \citep{chrpa2009reformulating}, and aims at removing useless actions from a ground planning task, to reduce the branching factor of a task at hand. Such useless actions are identified by analysing the relationship between predicates of the problem, and operators. In a nutshell, the idea is to change the representation of a planning task so that only potentially useful actions can be grounded and considered during search. 

In literature, two main classes of entanglements have been proposed: outer and inner. Outer entanglements \citep{chrpa2012exploiting,chrpa2018outer} identify useless actions on the basis of their relationships with predicates from the initial state or the goal state. 
Inner entanglements \citep{chrpa2019inner} focus instead on the relation between pairs of actions. There are two types of inner entanglements: \emph{entanglements by preceding}, which defines the case where a certain predicate is required by an operator as a precondition, and \emph{entanglements by succeeding}, which denotes the case where a particular operator makes a predicate available as one of its effects.

According to the results presented in \citep{chrpa2012reformulating}, exactly finding entanglements is as hard as solving the planning task itself. For this reason, most of the work in literature focuses on approximate approaches for identifying entanglement relations. To exemplify how entanglements can reformulate a given planning task, here we show how the Gripper domain can be modified by entanglements by init for the operator \texttt{pick}. The basis of entanglement in this case for the particular encoding is that it is useful to perform a \texttt{pick} action only on balls that are in the initial room. In any other occasion, the \texttt{pick} action should not be considered. The relationship is therefore captured by the \texttt{(at ?b ?r)} predicate, that indicates the position of a ball and is a precondition of the considered operator. To implement the entanglement, a new predicate $p^{\prime}$ \texttt{(at-ent ?b ?r)} should be created and added to the domain model, having the same arguments of \texttt{(at ?b ?r)}. The \texttt{pick} operator should be modified by adding $p^{\prime}$ as a precondition. In the initial state description of the problem, an additional $p^{\prime}$ predicate should be created for each existing \texttt{(at ?b ?r)}. This is similar in nature to the auxiliary predicate added to preconditions to account for unforeseen circumstances to address the qualification problem in service specifications~\citep{Baryannis2017,Baryannis2013} An excerpt of the modified domain and problem models is presented in Figure \ref{fig:outer_ent}.


\begin{figure}
\small
\begin{verbatim}
(:predicates ... (at-ent ?obj ?room))
...
(:action pick
       :parameters (?obj ?room ?gripper)
       :precondition  (and  (ball ?obj) (room ?room) (gripper ?gripper) (at ?obj ?room) 
       (at-ent ?obj ?room) (at-robby ?room) (free ?gripper))
       :effect (and (carry ?obj ?gripper) (not (at ?obj ?room) (not (free ?gripper))))
...)
....
(:init ... (at ball4 rooma) (at ball3 rooma) (at ball2 rooma) (at ball1 rooma) ... 
(at-ent ball4 rooma) (at-ent ball3 rooma) (at-ent ball2 rooma) (at-ent ball1 rooma))
...
\end{verbatim}
\caption{\label{fig:outer_ent} An excerpt of the modified Gripper planning task when the \texttt{pick} operator is entangled by init on the basis of the \texttt{at} predicate.}
\end{figure}

\subsection{Macro-Operators} \label{mac}

Macro-operators (macros) represent a well-known and well-studied technique for enhancing performance of planning engines. In a nutshell, macros encapsulate sequences of (primitive) planning operators. Macros are encoded as ordinary planning operators and, hence, they can be added into planning domain models. Macros, informally speaking, provide short-cuts in the state space and, consequently, planning engines can generate plans in a smaller number of steps. This comes at the cost of an increased branching factor, since macros often have many more instances than primitive operators and thus their use might introduce additional overheads as well as larger memory requirements.

The notion of macros can be traced back to 1970s and 1980s. REFLECT \citep{dawson1977role} builds macros from pairs of primitive operators that can be applied in sequence and share at least one argument. Macro Problem Solver \citep{korf1985macro} is capable of learning macros for achieving particular non-serialisable sub-goals (e.g. in Rubik’s cube). MORRIS \citep{minton1985selectively} learns macro-operators from parts of plans appearing frequently (S-macros) or being potentially useful despite having low priority (T-macros). FM \citep{mccluskey1987combining} learns sequences of operators frequently used together, and combines them in potentially long sequences called chunks. MACLEARN \citep{10.5555/1625135.1625259,iba1989heuristic} is a heuristic approach for learning re-usable macros for solving puzzle problems; the system includes three subcomponents that are in charge of proposing promising macros and testing their usefulness for solving problems. 

The first International Planning Competition, held in 1998, introduces PDDL that became the "de facto" standard language for planning models. The introduction of a language supported the design and testing of approaches for the generation and identification of macros, that started to thrive. MacroFF \citep{botea2005macro}, based on the well-known FF planning engine \citep{hoffmann2001ff}, generates macros according to a number of pre-defined rules (e.g., the ``locality rule'') that apply on adjacent actions in training plans. MacroFF is capable of generating planner-independent macros, that can be added to domain models as standard operators, or planner-specific macros for FF, that can be provided as additional input to the planning engine. DHG \citep{armano2004automatic} is able to learn macro-operators from static domain analysis by exploring a graph of dependencies between operators. WIZARD \citep{newton2007wizard} is a framework that exploits genetic programming to create macros; starting from the primitive operators of the domain model, WIZARD leverages genetic algorithms to combine them into useful macros for a given planning engine. \cite{chrpa2010generation} propose an approach for identifying suitable macros by looking at action dependencies in generated plans. \cite{dulac2013learning} propose to exploit n-gram algorithm to analyse training plans to automatically learn macros. 

DBMP/S \citep{hofmann2017initial} applies Map Reduce for learning macros from a large training plan databases. More recently, the same authors propose an approach for generating macros for ADL domain models, that includes PDDL features that are rarely supported by more traditional methods \citep{hofmann2020macro}. CAP \citep{asai2015solving} exploits component abstraction, that allows to cluster together similar objects (introduced by MacroFF), for generating sub-goal specific macros. In other words, CAP divides complex planning problems into independent sub-problems by abstracting the components of the original problem. Then it finds sub-plans for each sub-problem, and connects the actions of every sub-plan into a solo macro operator. BloMa \citep{chrpa2015exploiting} leverages block deordering, which rearranges plans into “blocks” that can no longer be deordered \citep{siddiqui2012block}, for generating longer macros. In particular, BloMa generates a large pool of macros from ``macroblocks'', which are derived from ``blocks'' by applying a set of rules. \cite{chrpa2019improving} introduce the idea of critical section macros, that are inspired by parallel computing critical sections; such macros aim at capturing a whole activity that deals with a limited resource (or more limited resources). Finally, \cite{castellanos2021era,castellanos2021relevance} introduce ERA, an approach for extracting macros from plans that is based on pattern mining; an important feature of ERA is its ability to identify macros even if the included operators are not always adjacent in the considered plans. 

A different line of work looks into exploiting the notion of entanglements to identify promising macros. \citep{chrpa2010combining} focuses on combining both macro-operators and entanglements in order to get the benefit behind each of them, as macros can reduce the size of the search space whereas the usage of entanglements is capable of reducing the branching factor which may occur because of the generated instances of macros.
\citep{chrpa2013generating} propose an automated approach that combine two primitive operators that are linked by inner entanglement relationships, and leverages on such relationships to eliminate one or both of the primitive operators from the domain model. MUM \citep{chrpa2014mum} is a learning system that exploits outer entanglements as heuristics in the process of generating macros. Macros generated by MUM have a limited number of instances, specifically, the number of macro instances has to be in the same order of magnitude as the number of primitive operator instances. OMA \citep{chrpa2015online} is capable of generating macros online, i.e. without the need for offline training, by considering entanglement relations between operators of the domain model.  

Notably, there is also a line of work that considers the problem of identifying the best (set of) macros to be used, given an initial pool of candidates. \cite{alhossaini2013instance} select problem-specific macros from a given pool of macros (hand-coded or generated by another technique) using a specifically trained predictor. ASAP \citep{vallati2013automatic} uses a set of provided training plans to identify the best combination of planning engine and macro set (also considering entanglements) to be used on a given domain. PbP \citep{gerevini2014planning} uses statistical tests to identify the most promising portfolio of planning engines and macro actions to be used for solving challenging planning instance. Finally, MeVo \citep{vallati2020mevo}, given a large pool of macros, can evolve over time the best set of macros to be used by a planning engine for solving a continuous stream of problems from a considered domain. This features allows MeVo to overcome the issue of having training instances that are not representative of the testing ones.



Technically speaking, the way in which a macro-operator is generated by assembling two operators is straightforward and quite similar to the way composition works in the case of services~\cite{Baryannis2014}. Considering two operators $a_{i}$ and $a_{j}$, the resulting macro encapsulating their execution in sequence can be generated as follows: 

$pre\left(a_{i, j}\right)=Pre\left(a_{i}\right) \cup\left(Pre\left(a_{j}\right) \backslash eff^{+}\left(a_{i}\right)\right)$

$eff^{-}\left(a_{i, j}\right)=\left(eff^{-}\left(a_{i}\right) \cup eff^{-}\left(a_{j}\right)\right) \backslash eff^{+}\left(a_{j}\right)$

$eff^{+}\left(a_{i, j}\right)=\left(eff^{+}\left(a_{i}\right) \cup eff^{+}\left(a_{j}\right)\right) \backslash eff^{-}\left(a_{j}\right)$\\


More than two operators can be encapsulated in a single macro by iteratively repeating the described process. Figure \ref{fig:fig_macros} shows the macro-operator \texttt{pick-move-drop} for the Gripper domain model. It has been generated by composing the primitive operators \texttt{pick}, \texttt{move}, and \texttt{drop}, with the idea of providing a single operator that can represent a whole movement of a ball from its initial position to its goal position. It can be added to the original domain model, and planning engines will take it into account when solving a given planning task. Notably, the resulting plans may include the macro operator and therefore, to be valid with regards to the original domain model, will need to be parsed. 

\begin{figure}[t]
\small
\begin{verbatim}
    (:action pick-move-drop  
       :parameters (?from ?to ?obj ?gripper) 
       :precondition (and  (ball ?obj) (gripper ?gripper) 
                (at ?obj ?from) (at-robby ?from) (free ?gripper)
                (room ?from) (room ?to)) 
       :effect (and  
            (not (at ?obj ?from))  
            (not (at-robby ?from)) 
            (at ?obj ?to) 
            (at-robby ?to)))	
\end{verbatim}    
    \caption{An example macro-operator that encapsulates the sequence of operators pick(?obj ?from ?gripper), move(?from ?to), drop(?obj ?to ?gripper). For the sake of clarity, we use the same name for matched variables between operators.}
    \label{fig:fig_macros}
\end{figure}

\subsection{Operators Elimination}
Operators elimination aims at reducing the branching factor by removing from the domain model those operators whose effects can be achieved by executing a sequence of other operators. In a nutshell, the main point is to minimise the size of the model by identifying operators that can be considered redundant. It should of course be noted that it may not always be possible to remove operators from a domain model. \cite{haslum2000planning} introduce a technique for performing operators elimination, by formally defining the notions of a redundant operator and of a minimal set of operators, also proposing a greedy algorithm to identify the minimal set of operators for a given domain model. 

Operators elimination may be of limited impact when used on its own, particularly in the case of highly engineered domain models. However, it can be particularly helpful when performed after another reformulation approach. For instance, \cite{chrpa2009reformulating} discuss operators elimination after the exploitation of entanglements, and there has been a large body of work investigating the elimination of primitive operators that are encapsulated into macros (e.g., \citep{chrpa2019improving,chrpa2010generation,alhossaini2013instance}). 


Considering the Gripper domain, in the original model shown in Figure \ref{fig:gripper_domain} there are no operators that are redundant and can be removed. However, if the domain model is extended by adding the macro shown in Figure \ref{fig:fig_macros}, then both \texttt{pick} and \texttt{drop} operators can be removed without compromising solvability of problems sharing the structure of the one shown in Figure \ref{fig:gripper_problem} -- i.e. where the goal is not for the robot to hold a ball in its gripper. 


\subsection{Bagged Representation}
In domain models encoded in PDDL, it is usually the case that each object is uniquely identified, even if it is not important to distinguish between objects. In the presence of large sets of objects, this can lead to an explosion of the combinatorial problem, that needs to take into account the specific information of each individual object. The bagged representation reformulation addresses the above criticism: in cases where only the number of objects is relevant, and it is not important to have the ability to distinguish between objects, they can be represented as ``bags'' of identical objects. The main advantage is to reduce the branching factor, by basically pruning states that are identical but for the specific object.

Bagged representation was first introduced in 2013 \citep{riddle2013problem}, and then further extended by providing an in-depth analysis of its impact on well-known benchmarks and automated techniques to perform the reformulation \citep{riddle2015bagged,riddle2015bagged2,riddle2015automated,riddle2016improving}. 



When it comes to reformulating the Gripper domain model, objects of type \texttt{ball} are an excellent candidate to be represented using bagged representation. It is important to know their number in every room, but it is not important to know which specific ball is where. Figure \ref{fig:bagged_domain} presents an excerpt of the reformulated domain and problem models. Objects of type ball are removed, and are substituted by counters. The predicate \texttt{count} is used to record the number of balls in a room. Predicate \texttt{more} is used to link together the different possible values of a counter. This is required because in classical planning there is no notion of numeric elements, so this allows to use Boolean predicates such as \texttt{more} as counters. The reformulated operator \texttt{pick} is also shown in Figure~\ref{fig:bagged_domain}, with the main change being that it now updates the counter of the room where it is applied. The \texttt{drop} operator (not shown in the figure) is modified in a similar way. Finally, the problem model is reformulated by removing the ball objects, appropriately setting the initial values of the counters for the considered rooms, and expressing the goal in terms of the number of balls that need to be in the final room.


\begin{figure}
\small
\begin{verbatim}
(:predicates ... (count ?b ?r ?n) (more ?n1 ?n2))
    
(:action pick 
    :parameters (?n1 ?n0 ?obj ?room ?gripper)
    :precondition (and (ball ?obj)(room ?room)
            (gripper ?gripper)(at-robby ?room)(free ?gripper)
            (more ?n1 ?n0)(count ?obj ?room ?n0))
    :effect (and (carry ?obj ?gripper)
        (not (count ?obj ?room ?n0))
        (count ?obj ?room ?n1)(not (free ?gripper))))
...)
....

(:objects n4 n3 n2 n1 n0 rooma roomb ballX left right)
(:init .... (ball ballX) (more n0 n1) (more n1 n2)(count ballX rooma n4)
(count ballX roomb n0) ...)
(:goal (count ballX roomb n4))
)
\end{verbatim}
    \caption{An excerpt of the Gripper domain and problem models reformulated using bagged representation for objects of the type \texttt{ball}.}
    \label{fig:bagged_domain}
\end{figure}

\subsection{Action Schema Splitting}

The presence of operators with a large number of parameters can be problematic for the grounding step of domain-independent planning engines, as they usually have to instantiate all possible actions before filtering out those that are irrelevant. The number of ground actions grows exponentially with the number of parameters and the number of objects of the problem to solve. In order to address this issue, the idea of action schema splitting~\citep{areces2014optimizing} is to split large operators into smaller ones (with regard to the number of parameters). The main aim is to break the exponential growth by dividing parameters between different operators. While the idea is intuitively easy, its exploitation has a number of hidden issues to be taken into account. When an operator is broken into two smaller operators, the order in which these two operators are executed becomes important, as is to consider whether their execution can be interleaved with different operators -- that can change the state of the world in some unexpected ways. 

Considering the example Gripper domain model, in the original version of the model there is no operator that is suitable to be split using the described methodology. However, action schema splitting can be understood also as the ``opposite'' of macro-operators and in fact, it can also be used to find reformulations of models by first encapsulating primitive operators into macros, and then splitting them in different ways. For this reason, to exemplify the use of this technique, we consider the macro encapsulating \texttt{pick, move, drop} operators shown in Figure \ref{fig:fig_macros} and we split it into two operators, \texttt{pick-move} and \texttt{drop}. Figure \ref{fig:schema_domain} shows the resulting operators. It is worth highlighting that the example also shows a drawback of the action schema splitting technique, i.e. the fact that the resulting operators may have the same number of parameters as the initial large operator.

\begin{figure}[h]
\small
\begin{verbatim}
   (:action drop
       :parameters  (?obj  ?room ?gripper)
       :precondition  (and  (ball ?obj) (room ?room) (gripper ?gripper)
			    (carry ?obj ?gripper) (at-robby ?room))
       :effect (and (at ?obj ?room) (free ?gripper) (not (carry ?obj ?gripper))))

    (:action pick-move
       :parameters (?obj ?from ?to ?gripper)
       :precondition  (and  (ball ?obj) (room ?from) (room ?to) (gripper ?gripper)
			    (at ?obj ?from) (at-robby ?from) (free ?gripper))
       :effect (and (carry ?obj ?gripper) (at-robby ?to) (not (at ?obj ?from))
		    (not (at-robby ?from)) (not (free ?gripper))))
\end{verbatim}   
    \caption{Action Schema Splitting applied to the macro operator shown in Figure \ref{fig:fig_macros}, and resulting in two operators.}
    \label{fig:schema_domain}
\end{figure}


\subsection{Domain Model Configuration}

It is well known that the way in which elements of planning models are ordered can have an impact on the performance of domain-independent planning engines \citep{howe2002critical}. In this context, the term elements can refer to operators, pre and post conditions, predicate definitions, etc. for the domain models; objects, initial and goal state predicates listing for problem definitions. 

The idea behind domain model configuration is to identify an ordering of elements that can improve the performance of a considered domain-independent planning engine. \cite{vallati2015effective} introduce an approach that leverages on algorithm configuration techniques to identify a suitable configuration of a domain model to improve the performance of a planning engine. The work has been subsequently extended \citep{vallati2021importance} to consider also cases where macro-operators have to be added to the domain model. \cite{vallati2017improving} describe a method for the online reordering of domain models by means of dedicated heuristics, based on aspects such as the number of preconditions, number of effects, etc. 
In a different line of work, \cite{vallati2018general} explore the configuration of planning problem models, by considering a structure called Planning Encoding Graph (PEG) \citep{serina2010kernel} to produce information that helps in creating the basis on which the reordering of elements should be done. In a nutshell, the PEG can provide information about how important some objects of the problem are based on their involvement in predicates of both initial and goal descriptions. This knowledge can then be exploited in the ordering of the predicates, according to the objects that they deal with.

Considering the guidelines in \citep{vallati2015effective}, there are no unpromising operators that we may prefer to put last in the Gripper domain model. However, following the introduced notion of \textit{directionality}, we may change the ordering of operators to \texttt{pick, move, drop}. In this way, the ordering of operators follow the expected typical ordering of corresponding actions in solution plans that has been demonstrated to be useful for a range of planning engines. 


\section{Qualitative Comparison}
\label{sec:qual}

Having completed an overview of the existing literature on domain- and planner-independent reformulation for classical planning, we are now able to qualitatively compare the considered techniques. In fact, we will now focus on the advantages and disadvantages of the reviewed techniques, with the aim of providing useful guidelines for planning experts and practitioners in the process of selecting a promising technique to be used to improve planning performance on a domain of interest. Table \ref{tab:comparison} gives an overview of the reviewed reformulation techniques in terms of their main advantages and the major potential drawbacks. 


\begin{table}[!t]
\footnotesize
\begin{center}
\caption{Qualitative comparison of the main strengths and weaknesses of the reviewed reformulation techniques.}
\label{tab:comparison}
\begin{tabular}{p{4cm}|p{4.5cm}|p{5cm}}
\hline

 \textbf{Reformulation Approach} & \textbf{Benefits} & \textbf{Drawbacks} \\
 \hline
 Macro-Operators  & - Reduce depth (reduce the transitions needed to reach goal states).

  & - Increase branching factor. 

- Increase of the ground size.\\
\hline
Entanglements  & 
- Reduce the branching factor.

  & - Potentially incomplete.  \\
\hline
 Actions Elimination & - Reduce the branching factor.
 & - Rarely applicable. 
 
 - Potentially incomplete. \\
 \hline
 Bagged Representation
 & - Reduce the branching factor. & - Only applicable when having indistinguishable numerable objects. \\
 \hline
 Action Schema Splitting
 & - Reduce the ground size of the problem. & - Potential increase of the branching factor. \\
 \hline
Domain Model Configuration & - Ease the exploration of the search space. & -  Can have limited impact if planners re-order elements internally. 

- Potential reduction of performance if wrong ordering is used. \\
 \hline
\end{tabular}
\end{center}

\end{table}

As indicated in Table \ref{tab:comparison}, most reformulation techniques aim at making the exploration of the search space easier by reducing the branching factor, reducing the number of steps needed to reach a goal state, or reducing the ground size of the problem to be solved. However, every technique has potential drawbacks to be weighed in when deciding whether to exploit it. Drawbacks range from the serious potential loss of completeness to a more shallow limited impact on performance in the worst case. The use of macros can boost the performance of a planning engine by reducing the number of steps needed to reach a goal state, at the cost of a potentially significant increase in terms of the branching factor and a larger ground problem. Entanglements aim at reducing the branching factor and the ground size of a problem by eliminating unpromising actions, but in doing so can remove some or all of the paths to goal states. In a similar fashion, action elimination directly reduces the ground size by eliminating operators deemed to be useless, but on its own it is rarely applicable to well-formed domain models. Further, if applied in an ``aggressive'' way, this may undermine completeness. Bagged representation provides an elegant way to reduce branching factor by removing the differences between objects, but it is again a technique that is rarely applicable on its own. Action schema splitting aims at reducing the ground size of a problem by splitting complex operators, but this comes at the cost of increasing the branching factor. Finally, configuring a domain model can help improve the performance of planning engines by listing the most important planning elements first, but if used inappropriately it can also lead to adverse effects.

It should be noted, however, that in many cases it is possible to combine different techniques to maximise performance boost and mitigate drawbacks of individual techniques \citep{vallati2020knowledge}. As mentioned in previous sections, macros are frequently combined with other reformulation techniques. For instance, they have been used with entanglements \citep{chrpa2014mum}, with action elimination \citep{chrpa2022}, or with domain model configuration \citep{vallati2021importance}. While domain model configuration can be straightforwardly combined with any other reformulation technique, the combination of macros and entanglements can lead to  significant performance improvements, as entanglements are tackling the main drawback that is associated with the use of macros, i.e. the branching factor.  

The main take-home messages that can be diluted by the performed systematic review of the literature, can be summarised as follows.
\begin{itemize}
    \item The majority of existing literature is dedicated to techniques for generating macros, while comparatively limited effort has been devoted to investigating alternative or different reformulation techniques. Admittedly, the planning community should aim at expanding the spectrum of reformulation techniques as much as possible, to foster their combination and the possibility to use them fruitfully in large and challenging domain models --where they are needed the most. 
    \item A substantial number of existing reformulation techniques aims at addressing issues that are typical of planning engines that reason on the basis of a ground representation. This focus is historically motivated, as domain-independent planning engines traditionally ground the lifted PDDL representation. However, as engines capable of reasoning with lifted or partially ground representation are gaining momentum (see for instance \citep{horvcik2021endomorphisms,correa2020lifted}), reformulations that are effective also on lifted models can and should be investigated more convincingly.
    \item On a similar note as the previous point, there is a lack of reformulations that look into exploiting the potential synergies with planning approaches that are based on compiling the planning problem into an equivalent SAT or ASP problem to be solved. Given the fact that reformulations for such languages have been proposed (see for instance \citep{dodaro_maratea_vallati_2022}), there may be changes in PDDL models that can result in performance boost in the resulting compiled instances. 
    \item A potentially interesting area to explore is the reformulation of problem models. Existing techniques tend to have a strong focus on the domain model, to be widely applicable on problems from the same domain. However, as planning is more and more used in real-world applications, the complexity of problems is increasing as well. Further, in some domains like urban traffic control \citep{DBLP:conf/aips/McCluskeyV17} or energy network balancing \citep{DBLP:conf/aips/PiacentiniAFL13}, large parts of the problem models remain the same between problems (i.e. the network description); so it would we worth exploring if problem models can benefit from a targeted reformulation. 
    \item We recognise that further work is needed to improve the usability of planning techniques in real-world applications, as the main focus of existing reformulation techniques is still on classical planning. However, as demonstrated by a number of recent works, in many cases classical planning reformulation approaches can be extended to, or provide inspiration for, non-classical systems. 
\end{itemize}

\section{Beyond classical planning} \label{sec:non}

In this review, we focused on reformulation techniques for classical planning. However, it is worth mentioning that there is also a (limited) body of work looking at reformulation for problems beyond classical planning. In this section we provide an overview, by no means complete, of some of the reformulation approaches introduced for planning problems beyond classical. On the one hand, non-classical planning models include additional language features that can complicate the reformulation process. Examples of languages that have the expressive power to represent non-classical prolems include PDDL 2 \citep{fox2003pddl2} and PDDL 3 \citep{gerevini2009deterministic}, for numeric, temporal planning and for encoding preferences; PDDL + \citep{fox2006modelling} for mixed discrete-continuous problems, PPDDL \citep{younes2004ppddl1} and RDDL \citep{sanner2010relational} for probabilistic planning, and MA-PDDL \citep{kovacs2012multi} for multi-agent problems.
On the other hand, it can be argued that non-classical planning is needed in most real-world applications, and it is therefore imperative to investigate techniques and approaches to boost planning performance in such circumstances. 

A line of research that looks into extending reformulation techniques for classical planning to more expressive cases is that of \cite{chrpa2015towards}, which  investigates the use of entanglements in numeric planning problems. Specifically, the authors extend the notion of outer entanglements to allow them to handle numeric variables. Similarly, \cite{scala2014plan} extended macros to be used in the presence of numeric fluents. Finally, \cite{DBLP:conf/iccS/FrancoVLM19}  present a technique to reduce the ground size of PDDL+ planning problems by reducing the arity of sparse predicates, drawing a parallel to bagged representation for classical planning. 

A different line of research focuses instead on translating a planning model from an original input language, to a different less-expressive one. The main advantage of this approach is increasing the number of planning engines that are able to reason upon the planning problem, and leverage existing robust technologies devised for solving more restricted cases. Examples of this class of reformulation approaches include \citep{percassi2021translations}, that translates PDDL+ problems into PDDL2.1 ones, \citep{grastien:20:aij,taig2013compiling} that provides approaches for translating of conformant planning problems into classical problems, the re-representation of uncertainty in conformant planning problems \cite{palacios2009compiling}, the translation of complex temporal aspects in PDDL2.1 \citep{5671413}, and the removal from PDDL3 of soft trajectory constraints \citep{DBLP:conf/aips/PercassiG19}.

\section{Conclusion}

Reformulation represents a well-known class of approaches for improving the performance of domain-independent planning systems. The main idea is to re-represent a given planning problem in a way that allows to increase the efficiency of a selected planning engine to be used to solve the considered problem or class of problems. In this paper, we reviewed the state of the art of reformulation approaches for classical planning. We first presented in detail the different techniques, and then took the opportunity to provide a qualitative comparison of their expected benefits and drawbacks, with the aim of providing some useful guidelines to select the most appropriate reformulation to be used for a considered problem. Notably, the provided comparison helps also in highlighting potentially fruitful combinations of reformulation techniques. Finally, to emphasise the importance of reformulation for planning models beyond classical, we briefly presented well-known work in the area.


\ack Mauro Vallati is supported by the UKRI Future Leaders Fellowship [grant number MR/T041196/1].

\bibliographystyle{abbrvnat} 
\bibliography{bibliography}

\end{document}